\documentclass[manuscript,screen]{acmart}

\AtBeginDocument{%
  \providecommand\BibTeX{{%
    \normalfont B\kern-0.5em{\scshape i\kern-0.25em b}\kern-0.8em\TeX}}}

\setcopyright{acmcopyright}
\copyrightyear{2021}
\acmYear{2021}
\acmDOI{10.48550/arXiv.2411.00190}

\acmConference[KDD '21]{KDD '21: ACM SIGKDD Conference on Knowledge Discovery and Data Mining, Workshop on Measures and Best Practices for Responsible AI}{August 14--18, 2021}{Singapore}
\acmBooktitle{KDD '21: ACM SIGKDD Conference on Knowledge Discovery and Data Mining, Workshop on Measures and Best Practices for Responsible AI, August 14--18, 2021, Singapore}
\acmISBN{978-1-4503-XXXX-X/18/06}

\begin{document}

\title{Monitoring fairness in machine learning models that predict patient mortality in the ICU}



\author{Tempest A. van Schaik}
\email{tempest.van@microsoft.com}
\affiliation{%
  \institution{Microsoft}
  \city{Redmond}
  \country{USA}
}

\author{Xinggang Liu}
\affiliation{%
  \institution{Philips}
  \city{Baltimore}
  \country{USA}}

\author{Louis Atallah}
\affiliation{%
 \institution{Philips}
 \city{Baltimore}
 \country{USA}}
 
 \author{Omar Badawi}
\affiliation{%
  \institution{Philips}
  \city{Baltimore}
  \country{USA}
}


\begin{abstract}

This work proposes a fairness monitoring approach for machine learning models that predict patient mortality in the ICU. We investigate how well models perform for patient groups with different race, sex and medical diagnoses.  We investigate \emph{Documentation bias} in clinical measurement, showing how fairness analysis provides a more detailed and insightful comparison of model performance than traditional accuracy metrics alone.

\end{abstract}


\maketitle

\section{Introduction}

Benchmarking against other Intensive Care Units (ICUs) can provide ICU staff and hospital managers with a broader view and clearer perspectives of targets for improvement. Benchmarking can include comparing an ICU's actual performance with predicted performance. The increased interoperability of medical devices, electronic health records (EHRs) and information systems has improved the acquisition and presentation of data to healthcare professionals. This data has enabled the training of predictive models. However, this plethora of data sources has also introduced new risks that societal bias will lead to machine learning systems with fairness issues for patient groups. In addition, when variations in data documentation are non-random, significant bias can be introduced, improving, or worsening measured performance for an institution relative to peers. This work focuses on ICU mortality benchmarking. In particular, we analyze the fairness of a model based on Generalised Additive Models (GAM) \cite{Hastie1990} that predicts mortality in the ICU. This model is used to compare actual versus predicted outcomes to assess ICU performance.

\section{Fairness Metrics in production}

We developed a new model \cite{LiuDocumentationBias} based on GAM , and compare this to an older, traditional model used for predictive analytics in the ICU. Both models are trained on the same clinical data \cite{Pollard2018} with the same features. However, for the purpose of fairness analysis, both models are a black box.

Machine learning systems have the potential to cause harm and lead to incorrect decision-making if they are unfair. In order to check whether our new ICU model makes predictions that are fair for different patient groups, we used the Fairlearn (version 0.4.6) Python package \cite{bird2020fairlearn}. Amongst the tools in the Fairlearn package are metrics for assessing the fairness of a model. Although technology alone cannot guarantee fairness, the Fairlearn package does help us check for negative impacts on groups of people, such as those defined in terms of sensitive features like race or sex. Specifically, we focus on quality of service: does the new model work as well for different patient groups? 

Removing sensitive features from the data we train our models on does not always guarantee fairness. For example, clinical models that are not trained with historical race data (i.e. race is not a feature in the training dataset) can still be racially biased \cite{ZiadObermeyerBrianPowers2019}. In healthcare, features like sex and race can sometimes affect outcomes. With Fairlearn, the sensitive features may or may not be features that the model was trained on. Although our model was not trained using race, Fairlearn allows us to assess how the model performs for different races.

Monitoring is an important aspect of Responsible AI because it provides visibility into how the machine learning system is working. In order to build reliable and maintainable machine learning models, we developed a monitoring system for: input data drift; training logs (how long each training step takes, accuracy on test data); accuracy of predictions; full traceability of model predictions, model code, random seeds, and training data; and, fairness metrics.

\subsection{Fairness analysis of sex, race and diagnosis}

In order to make use of the metrics available from Fairlearn, we designed a schema to create logical groupings of metrics and to be dynamically sized based on any number of sensitive features and any number of levels (i.e. possible values). The full schema is shown in the Appendix. These categories (e.g. binary sex) arise from upstream data capture and formatting and represent how the data is made available to researchers. Table~\ref{tab:sexrace} shows some of the schema for fairness metrics, including the (categorical) sex of patients (“catSex”). The Metric column contains accuracy and fairness metrics. Feature Level represents the different possible values within this Sensitive Feature. Threshold is the value used to convert a probability (mortality risk) between 0 and 1, into a binary value, for the metrics that need it. The overall accuracy, area under the Receiver Operating Characteristic curve (auROC), is 0.92331, while accuracy for Female and Non-Female patients is 0.92187 and 0.9245 respectively. This very small difference shows that the model behaves very similarly for Female and Non-Female patients. We chose auROC as an accuracy metric, but other accuracy metric(s) could also be appropriate. The schema presents performance metrics like accuracy in the same place as fairness metrics, to give them equal importance when viewed.
 
\begin{table}
\caption{A selection of fairness metrics for patient groups of different sex, race and medical diagnoses.}
\label{tab:sexrace}
\begin{tabular}{|c|c|c|c|c|}
\hline 
Metric & Value & Sensitive feature & Feature level & Threshold\tabularnewline
\hline 
\hline 
auROC & 0.92331276 &  &  & \tabularnewline
\hline 
auROC by group & 0.921873233 & catSex & Female & \tabularnewline
\hline 
auROC by group & 0.924500959 & catSex & Non-Female & \tabularnewline
\hline 
auROC by group & 0.921579487 & race & Caucasian & \tabularnewline
\hline 
auROC by group & 0.933864299 & race & Native American & \tabularnewline
\hline 
min auROC over groups & 0.816517641 & dxGroup &  & \tabularnewline
\hline 
max auROC over groups & 0.962784631 & dxGroup &  & \tabularnewline
\hline 
auROC by group & 0.816517641 & dxGroup & CardiacArrest & \tabularnewline
\hline 
auROC by group & 0.82984737 & dxGroup & ARDS & \tabularnewline
\hline 
auROC by group & 0.962784631 & dxGroup & DKA & \tabularnewline
\hline 
selection rate & 0.203554374 &  &  & 0.05\tabularnewline
\hline 
demographic parity difference & 0.849936685 & dxGroup &  & 0.05\tabularnewline
\hline 
selection rate by group & 0.855637145 & dxGroup & CardiacArrest & 0.05\tabularnewline
\hline 
selection rate by group & 0.010640763 & dxGroup & DKA & 0.05\tabularnewline
\hline 
\end{tabular}

\end{table}

Performing the fairness analysis for different race groups, with auROC ranging around 0.92 (lowest) to 0.93 (highest) for different races.
 
We then extended our analysis to explore additional model features, like diagnosis group ("dxGroup"). This is the diagnosis that the patient received on admission to the ICU. Unlike sex and race, diagnosis group has a large difference in auROC in Table~\ref{tab:sexrace}, with 0.82 being the lowest and 0.96 being the highest.
 
This may be because patients who come into the ICU with certain diagnosis groups, such as Acute Respiratory Distress Syndrome (ARDS) have lower accuracy predictions. This could be due to ARDS typically occurring in people who are already critically ill or who have significant injuries so it would be hard to predict their recovery patterns. In contrast, recovery patterns are easier to predict for other patients, such as those with Diabetic Ketoacidosis (DKA). These patients, who may be young and otherwise healthy can be having an issue managing their blood sugar. They are likely to make a more predictable recovery once they receive a treatment like insulin. 

Next we consider selection rate, which is the fraction of predictions matching the mortality outcome. Mortality is predicted for 0.204 of the ICU stays, when we choose a prediction threshold of 0.05 (different thresholds could be chosen). 

A high demographic parity difference (close to 1) tells us that there is a large discrepancy in selection rate across diagnosis groups. Demographic parity difference for diagnosis group, i.e. the difference between the highest and lowest selection rate across different diagnosis groups is 0.86. When we sort the diagnosis groups by selection rate, we see that cardiac arrest has a very high selection rate. This tells us that the model is frequently predicting mortality for patients who come into the ICU with cardiac arrest, which is a very serious condition. Diabetic Ketoacidosis (DKA) appears again, this time with a very low selection rate. This could be due to the clear treatment path (e.g. insulin) that leads to good patient outcomes. There may also be a low sample size of patients in some categories, which may lead to decreased model accuracy in less prevalent diagnosis groups.

\section{Documentation bias}

Next, we extend fairness analysis to address the issue of \emph{documentation bias}. We define \emph{documentation bias} as the measurement error in clinical features that are used to train a model. We focus on the Glasgow Coma Scale (GCS), which is a standard assessment of the level of consciousness, where points are summed for eye opening, verbal and motor response. A patient can score a minimum of 3 points (completely unresponsive) and a maximum of 15 points (responsive) (see Appendix). 
 
GCS is an important feature for predicting mortality, as a low score indicates that a patient is very unwell. Sometimes the GCS cannot be accurately determined, for example when a patient is heavily sedated. In this case, a healthcare professional should capture the null data as “Unable to score due to medications”. However, there is a lack of clear guidance for this situation \cite{Laureys2005}, and sometimes it is captured as the minimum (GCS=3) or the maximum (GCS=15). A GCS=3 score indicates that the patient is in a coma, so using it when the patient is healthier than that creates measurement error. Different ICUs may have different documentation norms, and they may change their documentation without notice and without recording what has changed. This systemic bias in the capture of GCS decreases model accuracy, negatively affecting the benchmark reports which are created for ICUs.

A model should perform equally well for a patient whether they visit one ICU or another, regardless of the hospital’s documentation norms. The prediction for a patient should not depend on how their healthcare professional captured GCS data, which we consider a fairness issue for patients. We compared robustness to documentation bias of the new GAM-based model to the old model. In order to do this, we categorized ICUs by how often GCS=3 is captured: high amount of GCS=3 (top 5\%), low amount of GCS=3 (bottom 5\%), medium amount (5th to 95th percentile). Thus, each patient stay is in an ICU of one of these three categories. We then used this new feature as a sensitive feature to perform fairness analysis on. First, we see results for an old model in Table~\ref{tab:GCS_models}.

\begin{table}
\caption{Comparison of fairness metrics in the old model, and in the new GAM-based model designed to be robust to documentation bias}
\label{tab:GCS_models}
\begin{tabular}{|c|c|c|c|c|c|}
\hline 
Metric & Value (old model) & Value (GAM-based model)) & Sensitive feature & Feature level & Threshold\tabularnewline
\hline 
\hline 
auROC & 0.882443824 & 0.92331276 &  &  & \tabularnewline
\hline 
equalized odds ratio & 0.432985061 & 0.552091966 & GCS3 &  & 0.05\tabularnewline
\hline 
auROC by group & 0.890022433 & 0.929648363 & GCS3 & highGCS3 & \tabularnewline
\hline 
auROC by group & 0.851896713 & 0.910848335 & GCS3 & lowGCS3 & \tabularnewline
\hline 
auROC by group & 0.881876847 & 0.922596613 & GCS3 & medGCS3 & \tabularnewline
\hline 
true positive rate by group & 0.915137615 & 0.890235911 & GCS3 & highGCS3 & 0.05\tabularnewline
\hline 
true positive rate by group & 0.687195122 & 0.722560976 & GCS3 & lowGCS3 & 0.05\tabularnewline
\hline 
true positive rate by group & 0.864222598 & 0.854690475 & GCS3 & medGCS3 & 0.05\tabularnewline
\hline 
false positive rate by group & 0.356012798 & 0.178050553 & GCS3 & highGCS3 & 0.05\tabularnewline
\hline 
false positive rate by group & 0.154148223 & 0.09830028 & GCS3 & lowGCS3 & 0.05\tabularnewline
\hline 
false positive rate by group & 0.27706954 & 0.166885698 & GCS3 & medGCS3 & 0.05\tabularnewline
\hline 
\end{tabular}
\end{table}
 
Results for the old model shows a low equalized odds ratio of 0.432985 (ideally it would be 1). This is the highest ratio between false positive rates per group, or true positive rates per group. This indicates a large discrepancy between groups when it comes to false positive rate. This is apparent when we consider ICUs that have a high amount of GCS=3 (“highGCS3”) which have a false positive rate of 0.356013. The model is likely to make many mortality predictions for this group given the poor GCS score. However, many of these predictions are false positives because patients are not really in a coma, they were just unable to be assessed due to sedation medications.

Next, we consider the GAM-based model, which has been designed to be more robust to measurement error and the resulting documentation bias. The GAM-based model shows in Table~\ref{tab:GCS_models} a lower false positive rate in ICUs that frequently document GCS=3. The equalized odds ratio is now higher (0.55209). Now we see that false positive rate in the “highGCS3” group is approximately halved (0.17805).

This is an encouraging result which suggests that the GAM-based model is not only more accurate overall (higher auROC) but also more robust to documentation bias (false positive rate for ICUs using GCS=3 for null data is halved). Traditionally we might have just compare two models by comparing their accuracy. However, the fairness analysis framework provides us with more informative, detailed and nuanced information for model comparison.

Other clinical features can also contain documentation bias. For example, the clinical diagnosis of sepsis when a patient arrives in ICU. There is some subjectivity in diagnosing sepsis \cite{Gyawali2019} as the definition is broad and medical guidelines are evolving. Therefore, it is important to have predictive models that are robust to documentation bias, and our analysis shows how we can check them for disparities.

\section{Discussion}

Responsible AI is a socio-technical challenge and we cannot guarantee that a machine learning system will behave fairly for all users. Using fairness metrics to analyze model predictions does not guarantee that a dataset is without flaws. But, if we calculate fairness metrics for groups of patients as part of routine system monitoring, it does give us visibility into fairness issues. 

If disparities between patient groups are found during fairness analysis, they can be discussed with domain experts (ICU clinicians) to understand the underlying cause, whether clinical or due to societal bias. If fairness issues are found, then unfairness mitigation techniques could be explored (potentially with Fairlearn). Future improvements to the fairness metrics schema could include showing a sample size for each fairness metric, and showing a zero or one as an ideal value, to make the metrics easier to interpret. 

\section{Conclusion}

Fairness is especially important for machine learning models used in the medical domain. We explored the fairness of ICU mortality models for different patient groups (sex, race, and diagnosis). Fairness analysis helped us address the documentation bias in clinical data that could lead to decreasing model accuracy and can negatively affect benchmarking between different ICU units/hospitals. We showed that fairness metrics provide additional insights when comparing model performance, and can use them to show the value of an important new model which will enhance benchmark reports.

\bibliographystyle{ACM-Reference-Format}
\bibliography{references.bib}

\newpage
\section{Appendix}
\appendix

\begin{table} [b]
\caption{Fairness schema for assessing performance of an ICU model for different patient sex and race. }
\begin{tabular}{|c|c|c|c|c|}
\hline 
Metric & Value & Sensitive feature & Feature level & Threshold\tabularnewline
\hline 
\hline 
area under ROC & 0.92331276 &  &  & \tabularnewline
\hline 
selection rate & 0.203554374 &  &  & 0.05\tabularnewline
\hline 
mean prediction & 0.056617797 &  &  & \tabularnewline
\hline 
false negative rate & 0.144665557 &  &  & 0.05\tabularnewline
\hline 
false positive rate & 0.165409626 &  &  & 0.05\tabularnewline
\hline 
true negative rate & 0.834590374 &  &  & 0.05\tabularnewline
\hline 
true positive rate & 0.855334443 &  &  & 0.05\tabularnewline
\hline 
min auROC over groups & 0.921873233 & catSex &  & \tabularnewline
\hline 
max auROC over groups & 0.924500959 & catSex &  & \tabularnewline
\hline 
difference in auROC & 0.002627727 & catSex &  & \tabularnewline
\hline 
ratio in auROC & 0.997157681 & catSex &  & \tabularnewline
\hline 
demographic parity ratio & 0.979531041 & catSex &  & 0.05\tabularnewline
\hline 
demographic parity difference & 0.004213659 & catSex &  & 0.05\tabularnewline
\hline 
equalized odds difference & 0.004567504 & catSex &  & 0.05\tabularnewline
\hline 
equalized odds ratio & 0.976040592 & catSex &  & 0.05\tabularnewline
\hline 
auROC by group & 0.921873233 & catSex & Female & \tabularnewline
\hline 
auROC by group & 0.924500959 & catSex & Non-Female & \tabularnewline
\hline 
true positive rate by group & 0.852859451 & catSex & Female & 0.05\tabularnewline
\hline 
true positive rate by group & 0.857426954 & catSex & Non-Female & 0.05\tabularnewline
\hline 
false positive rate by group & 0.167604207 & catSex & Female & 0.05\tabularnewline
\hline 
false positive rate by group & 0.163588509 & catSex & Non-Female & 0.05\tabularnewline
\hline 
selection rate by group & 0.205856063 & catSex & Female & 0.05\tabularnewline
\hline 
selection rate by group & 0.201642404 & catSex & Non-Female & 0.05\tabularnewline
\hline 
min auROC over groups & 0.921579487 & race &  & \tabularnewline
\hline 
max auROC over groups & 0.933864299 & race &  & \tabularnewline
\hline 
difference in auROC & 0.012284812 & race &  & \tabularnewline
\hline 
ratio in auROC & 0.986845185 & race &  & \tabularnewline
\hline 
demographic parity ratio & 0.847799569 & race &  & 0.05\tabularnewline
\hline 
demographic parity difference & 0.032816044 & race &  & 0.05\tabularnewline
\hline 
equalized odds difference & 0.030529757 & race &  & 0.05\tabularnewline
\hline 
equalized odds ratio & 0.83323635 & race &  & 0.05\tabularnewline
\hline 
auROC by group & 0.927861192 & race & African American & 0.05\tabularnewline
\hline 
auROC by group & 0.923153043 & race & Asian & 0.05\tabularnewline
\hline 
auROC by group & 0.921579487 & race & Caucasian & 0.05\tabularnewline
\hline 
auROC by group & 0.928151561 & race & Hispanic & 0.05\tabularnewline
\hline 
auROC by group & 0.933864299 & race & Native American & 0.05\tabularnewline
\hline 
auROC by group & 0.932336779 & race & Other/Unknown & 0.05\tabularnewline
\hline 
true positive rate by group & 0.87477465 & race & African American & 0.05\tabularnewline
\hline 
true positive rate by group & 0.867370008 & race & Asian & 0.05\tabularnewline
\hline 
true positive rate by group & 0.850323261 & race & Caucasian & 0.05\tabularnewline
\hline 
true positive rate by group & 0.849474912 & race & Hispanic & 0.05\tabularnewline
\hline 
true positive rate by group & 0.868558626 & race & Native American & 0.05\tabularnewline
\hline 
true positive rate by group & 0.880004669 & race & Other/Unknown & 0.05\tabularnewline
\hline 
false positive rate by group & 0.175014258 & race & African American & 0.05\tabularnewline
\hline 
false positive rate by group & 0.17440808 & race & Asian & 0.05\tabularnewline
\hline 
false positive rate by group & 0.164947374 & race & Caucasian & 0.05\tabularnewline
\hline 
false positive rate by group & 0.145828241 & race & Hispanic & 0.05\tabularnewline
\hline 
false positive rate by group & 0.161726224 & race & Native American & 0.05\tabularnewline
\hline 
false positive rate by group & 0.164183726 & race & Other/Unknown & 0.05\tabularnewline
\hline 
selection rate by group & 0.213157424 & race & African American & 0.05\tabularnewline
\hline 
selection rate by group & 0.215610718 & race & Asian & 0.05\tabularnewline
\hline 
selection rate by group & 0.202881338 & race & Caucasian & 0.05\tabularnewline
\hline 
selection rate by group & 0.182794674 & race & Hispanic & 0.05\tabularnewline
\hline 
selection rate by group & 0.200121704 & race & Native American & 0.05\tabularnewline
\hline 
selection rate by group & 0.205774239 & race & Other/Unknown & 0.05\tabularnewline
\hline 
\end{tabular}
\end{table}

\begin{table}
\caption{Glasgow Coma Scale}
\begin{tabular}{|c|c|c|}
\hline 
Area Assessed & Response & Points\tabularnewline
\hline 
\hline 
Eye opening & Open spontaneously & 4\tabularnewline
\hline 
 & Open to verbal command & 3\tabularnewline
\hline 
 & Open in response to pain applied to the limbs or sternum & 2\tabularnewline
\hline 
 & None & 1\tabularnewline
\hline 
Verbal & Oriented & 5\tabularnewline
\hline 
 & Disoriented, but able to answer questions & 4\tabularnewline
\hline 
 & Inappropriate answers to questions; words discernible & 3\tabularnewline
\hline 
 & Incomprehensible speech & 2\tabularnewline
\hline 
 & None & 1\tabularnewline
\hline 
Motor & Obeys commands & 6\tabularnewline
\hline 
 & Responds to pain with purposeful movement & 5\tabularnewline
\hline 
 & Withdraws from pain stimuli & 4\tabularnewline
\hline 
 & Responds to pain with abnormal flexion (decorticate posture) & 3\tabularnewline
\hline 
 & Responds to pain with abnormal (rigid) extension (decerebrate posture) & 2\tabularnewline
\hline 
 & None & 1\tabularnewline
\hline 
\end{tabular}
\end{table}

\end{document}